\documentclass[10pt,journal,compsoc]{IEEEtran}
\usepackage[backend=bibtex,style=ieee,natbib=true,mincitenames=1]{biblatex} 
\ifCLASSINFOpdf
\else
\fi

\usepackage{booktabs} 
\usepackage{graphicx}   
\usepackage{epstopdf}
\usepackage{amsmath}
\usepackage[linesnumbered,boxed,ruled,vlined]{algorithm2e}
\usepackage{multirow}
\usepackage{xcolor}

\usepackage{wrapfig}
\usepackage{amssymb}
\usepackage{mathtools}
\usepackage{amsthm}
\usepackage{enumitem}
\usepackage{subcaption}

\newtheorem{theorem}{Theorem}[section]

\newtheorem{definition}[theorem]{Definition}

\begin{document}
%
\title{Attacking and Securing Community Detection: A Game-Theoretic Framework}
%
%
%
%

\author{Yifan~Niu,
        Aochuan~Chen,
        Tingyang~Xu,
        Jia~Li
\IEEEcompsocitemizethanks{\IEEEcompsocthanksitem Yifan Niu, Aochuan Chen, and Jia Li are with The Hong Kong University of Science and Technology. Jia Li is the corresponding author.\protect\\
E-mail: \{yniu669, achen149\}@connect.hkust-gz.edu.cn, jialee@ust.hk
\IEEEcompsocthanksitem Tingyang Xu is with DAMO Academy, Alibaba Group and Hupan Lab. \\
E-mail: xuty\_007@hotmail.com
}
}

%
%

\markboth{Journal of \LaTeX\ Class Files,~Vol.~14, No.~8, August~2015}%
{Shell \MakeLowercase{\textit{et al.}}: Bare Demo of IEEEtran.cls for Computer Society Journals}
%



\IEEEtitleabstractindextext{%
\begin{abstract}

It has been demonstrated that adversarial graphs, i.e., graphs with imperceptible perturbations, can cause deep graph models to fail on classification tasks.  In this work, we extend the concept of adversarial graphs to the community detection problem,  which is more challenging. We propose novel attack and defense techniques for community detection problem, with the objective of hiding targeted individuals from detection models and enhancing the robustness of community detection models, respectively. These techniques have many applications in real-world scenarios, for example, protecting personal privacy in social networks and understanding camouflage patterns in transaction networks.  To simulate interactive attack and defense behaviors, we further propose a game-theoretic framework, called CD-GAME. One player is a graph attacker, while the other player is a Rayleigh Quotient defender. The CD-GAME models the mutual influence and feedback mechanisms between the attacker and the defender, revealing the dynamic evolutionary process of the game. Both players dynamically update their strategies until they reach the Nash equilibrium. Extensive experiments demonstrate the effectiveness of our proposed attack and defense methods, and both outperform existing baselines by a significant margin.  Furthermore, CD-GAME provides valuable insights for understanding interactive attack and defense scenarios in community detection problems. We found that in traditional single-step attack or defense, attacker tends to employ strategies that are most effective, but are easily detected and countered by defender. When the interactive game reaches a Nash equilibrium, attacker adopts more imperceptible strategies that can still achieve satisfactory attack effectiveness even after defense.
\end{abstract}

}

\maketitle

\IEEEdisplaynontitleabstractindextext

%
\IEEEpeerreviewmaketitle

\IEEEraisesectionheading{\section{Introduction}\label{sec:introduction}}

%
%
%
%
\IEEEPARstart{C}{Community} detection is one of the most widely studied topics in the graph domain, which aims to discover groups of nodes in a graph such that the intragroup connections are denser than the intergroup ones \cite{wang2015community}. 
This technique has numerous real-world applications, including functional module identification in protein-protein interaction networks \cite{ahn2010link}, discovery of scientific disciplines in coauthor networks \cite{zhou2009graph}, and detection of fraudulent organizations in user-user transaction networks \cite{akoglu2015graph}. However, with the rapid development of community detection algorithms, people realize that their privacy is over-mined \cite{Chen2018GABasedQO}. In this context, some studies have started to explore techniques for hiding individuals or communities \cite{waniek2018hiding,fionda2017community}, as well as methods to degrade the overall performance of community detection algorithms \cite{Chen2018GABasedQO}, primarily using heuristics or genetic algorithms. Meanwhile, another line of research has emerged that focuses on enhancing the robustness of community detection models to mitigate the influence of adversarial perturbations, particularly using the randomized smoothing technique \cite{cohen2019certified}.

Recently, deep graph learning models \cite{perozzi2014deepwalk,kipf2017semi} have achieved remarkable performance in many graph learning tasks.  Meanwhile, some studies \cite{zugner2018adversarial,dai2018adversarial} notice that deep graph models can be easily attacked on tasks such as node/graph classification.  Motivated by these findings, in this work, we aim to address the following question in community detection: (1) \emph{how vulnerable are graph learning based community detection methods? (2) Can we hide individuals by imperceptible graph perturbations?  (3) How can we construct a robust detection model? (4) How do interactive attack and defense evolve over time?} A good solution to this problem can benefit many applications, for example, the protection of personal privacy and the understanding of fraud escape.

\begin{figure*}
\begin{center}
\includegraphics [width=1\textwidth]{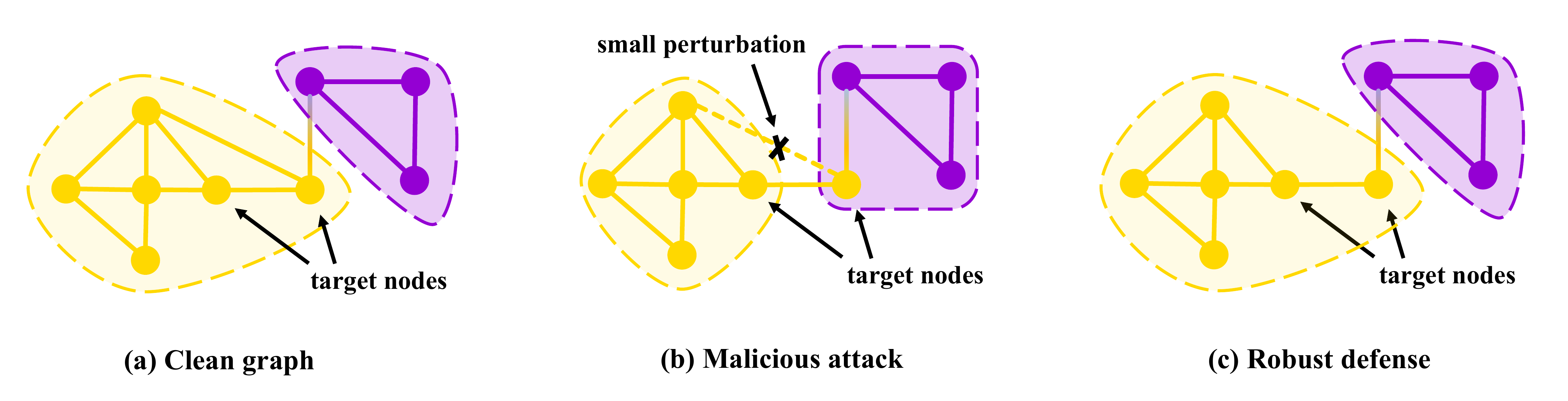}
\end{center}
\vspace{-0.5cm}
\caption{An example of adversarial attack and defense in community detection is illustrated. Initially, the two targets are clustered into one community (Fig.\ 1(a)). With a small malicious perturbation, such as deleting an edge, the two targets are assigned to different communities by the same detection method (Fig.\ 1(b)). To counter this, we develop a robust defense model that is able to maintain the original results against malicious attacks (Fig.\ 1(c)).}
\label{fig.example1}
\end{figure*}

Unlike adversarial attacks on node/graph classification where gradients \cite{zugner2018adversarial} or limited binary responses \cite{dai2018adversarial} of the target classifier are available, one challenge we face is that there is no feedback from the target model. For example, on social networks like Facebook or Twitter, community detection algorithms serve as the back-end for other purposes, such as advertising, which prevents direct interactions between the target model and individuals. To address this challenge, we designed a surrogate community detection model based on famous graph neural networks (GNNs) \cite{kipf2017semi} and a popular community detection measure \emph{normalized cut} \cite{shi2000normalized}.  We attack this surrogate community detection model and verify that the attack is able to be transferred to other popular graph learning-based community detection models.

For adversarial attacks, most existing methods address imperceptible perturbations employing either a discrete budget to constrain the number of allowable changes \cite{sun2018adversarial, dai2018adversarial} or a predefined distribution, such as a power-law distribution \cite{zugner2018adversarial}.  
One of the crucial challenges in generating an adversarial graph is the huge search space of candidate edges/nodes. Therefore, it is non-trivial to develop a generic solution that scales to large-size graphs.  Existing works, such as \cite{waniek2018hiding}, rely on heuristics to bypass this problem. However, they fail in finding optimal choices, especially for attributed graphs. In this work, we design a novel attack method, CD-ATTACK, which reformulates the complicated discrete combinatorial optimization problem into a locally differentiable problem. We observe that CD-ATTACK not only generates effective adversarial graphs subject to a given budget, but also demonstrates promising scalability.

Another line of research is for the opposite purpose, i.e. improving the robustness of community detection models.  \cite{jia2020certified} improves robustness of detection model with randomized smoothing \cite{cohen2019certified}. However, the proposed method falls short in detection accuracy, as it simply treats the graph structure as a set of edges and introduces random noises. Inspired by the fact that malicious attacks tend to destroy this similarity by adding or deleting connections between nodes from different or the same communities \cite{wu2019adversarial}, we propose a novel defense method, CD-DEFENSE, to improve the robustness of the detection model by reducing the high-frequency energy of the social network. Extensive experiments validate that CD-DEFENSE is effective in improving the robustness of community detection models.

Furthermore, an open research question is how to model the interactive attack and defense behaviors in community detection problems. Unlike existing works that focus on single-step attack or defense, a more complicated sequential attack-defense game has not been explored. In order to bridge the gap, we raise the following questions: \emph{How does this security game evolve? Does it eventually reach an equilibrium state?} We propose a game-theoretic framework, CD-GAME, to model this attack-defense game. The CD-GAME leverages mutual influence and builds feedback mechanisms between the attacker and the defender, revealing the dynamic evolutionary process of the game. In general, CD-GAME provides valuable insights for understanding the interactive attack and defense in community detection problem. We found that in traditional single-step attack or defense, attacker tends to employ strategies that are most effective, but are easily detected and countered by defender. When the interactive game reaches a Nash equilibrium, attacker adopts more imperceptible strategies that can still achieve satisfactory attack effectiveness even after defense.

Our contributions are summarized as follows.

\begin{itemize}
\item We propose a new graph learning based community detection model which relies on the widely used GNNs and the popular measure \emph{normalized cut}. It serves as the surrogate model to attack; furthermore, it can also be used for solving general unsupervised non-overlapping community detection problems.

\item We study adversarial attacks on graph learning-based community detection models by hiding a set of nodes. Our proposed CD-ATTACK reformulates this into a locally differentiable problem, and achieves superior attack performance with promising scalability.

\item We propose a novel defense method, CD-DEFENSE, to enhance robustness of detection model by reducing the high-frequency energy of the social network. CD-DEFENSE achieves superior defense performance compared to all competitors.

\item We propose a game-theoretic framework, CD-GAME, which provides valuable insights for understanding the interactive attack and defense scenarios.

\end{itemize}

The remainder of this paper is organized as follows.  Section \ref{def} introduces the problem definition. Section \ref{alt} describes the design of the CD-ATTACK attachment method, the CD-DEFENSE defense technique, and the CD-GAME game-theoretic framework. We report the experimental results in Section \ref{sec.exp} and discuss related work in Section \ref{sec:relatedwork}.  Finally, Section \ref{sec.con} concludes the paper.

\section{Problem Definition}\label{def}
We denote a set of nodes as $V=\{v_1, v_2, \ldots, v_N\}$ that represent entities, e.g., authors in a coauthor network, users in a user-user transaction network.  We use an adjacency matrix $A\in \mathbb{R}^{N\times N}$ to describe the connections between the nodes in $V$.  $A_{ij}\in \{0, 1\}$ represents whether there is an undirected edge between nodes $v_i$ and $v_j$ or not, for example, a coauthored article linking two authors, a transaction connecting two users.  In this study, we focus on an undirected graph; yet our method is also applicable to directed graphs.  We use $X=\{x_1, x_2, \ldots, x_N\}$ to denote the attribute values of the nodes in $V$, where $x_i\in \mathbb{R}^{d}$ is a vector of dimensions $d$.

The community detection problem aims to partition a graph $G = (V, A, X)$ into $K$ disjoint subgraphs $G_i = (V_i, A_i, X_i)$, $i=1, \ldots, K$, where $V = \cup_{i=1}^KV_i$ and $V_i\cap V_j = \emptyset$ for $i\neq j$.  In our study, we adopt this formulation, which produces non-overlapping communities.

In the context of community detection, we are interested in a set of individuals $C^+ \subseteq V$ who actively want to escape detection as a community or as a part of a community. For example, users involved in some underground transactions who are eager to hide their identity, or malicious users who intend to fool a risk management system.  Given a community detection algorithm $f(\cdot)$, the \emph{community detection adversarial attack} problem is defined as learning an attacker  to perform small perturbations on $G = (V, A, X)$, leading to $\hat{G} = (\hat{V},\hat{A},\hat{X})$, such that
\begin{align}\label{eq1}
\begin{split}
  \min \mathcal{L}(f(\hat{G}), C^+) \ \ \text{s.t.} \ \ Q(G, \hat{G}) < \epsilon,
\end{split}
\end{align}
where $\mathcal{L}(\cdot, \cdot)$ measures the quality of community detection results $f(\hat{G})$ related to the target $C^+$, and $Q(G, \hat{G}) < \epsilon$ is used to ensure imperceptible perturbations.  In this work, we focus on edge perturbations, such as edge insertion and deletion (i.e., $\hat{G} = (\hat{V}, \hat{A}, \hat{X}) = (V, \hat{A}, X)$).  Intuitively, we want to maximize the decrease in community detection performance related to the target set $C^+$ by injecting small perturbations.

Figure \ref{fig.example1}(b) illustrates an example of community detection adversarial attack in a user-user transaction network.  Initially, the two target individuals are clustered in a yellow community, representing a money laundering group.  One target, as a member of the community, would be suspected of money laundering given that the other is exposed somehow.  However, with a small perturbation by deleting an edge, the target individuals are assigned to different communities.  Thus, the two targets decrease the probability of being detected as a whole.

 As for enhancing the robustness of community detection, it is defined as learning a robust community detection model $f^{\ast}$, in which the detection results of $f^{\ast}(\hat{G})$ do not drift significantly from $f(G)$:
\begin{align}
\begin{split}
  \min \  | f^{\ast}(\hat{G}) - f(G) | \ \
  \text{s.t.}\ \ \  Q(G, \hat{G}) < \epsilon.
\end{split}
\end{align}
In essence, we want to maintain the clustering results of the community detection model against adversarial perturbations. Figure \ref{fig.example1}(c) illustrates an example of adversarial defense. Despite a small perturbation caused by the removal of an edge, the robust community detection model still successfully groups criminals into one community. 

\textbf{The Scope of Malicious Attacks.} Generally speaking, attackers are usually restricted to manipulating edges within their immediate neighborhood. They merely have the power to add or delete edges within a bounded-hop space. In our analysis, for any target $t \in C^+$, we assume that it can only apply adversarial perturbations within its $k$-hop neighborhood $\mathcal{N}_k(t)$.

\section{Methodology}\label{alt}
\subsection{Overall Framework}
In our problem setting, we propose a game-theoretic framework called CD-GAME, which consists of two modules: (1) an adversarial attacker, CD-ATTACK,  that generates attack graph $\hat{G}$ to perform \emph{unnoticeable} perturbations on the original graph $G$ in order to hide a set of individuals, and (2) a defender, CD-DEFENSE, that generates smooth social network $\bar{G}$ from $\hat{G}$ to mitigate the negative influence of adversarial perturbations and preserve the original detection result.  As we focus on the black box setting, which means that there is no specific community detection algorithm at hand, we need to instantiate a surrogate community detection model $f(\cdot)$ that can partition a graph into several communities in an unsupervised manner. 
These modules are highly interrelated. The adversarial attacker needs the feedback from $f(\cdot)$ to check if the goal of hiding is achieved or not, and the defender needs to enhance the robustness of the community detection algorithm $f(\cdot)$  w.r.t adversarial examples. However, when instantiating the two modules, we face three challenges as follows:

\noindent\textbf{Constrained attack graph generation}.  How to efficiently generate effective adversarial graphs that meet the imperceptibility requirement?

\noindent\textbf{Robust defense mechanism}. How to design a robust defense mechanism without being aware of the accurate attack location?

\noindent\textbf{Dynamic Game Framework}. How to effectively model the adversarial game process and consider the interaction between the two players?

These three issues make our problem very complicated.  In the following, we propose CD-ATTACK, CD-DEFENSE, and CD-GAME to address the three challenges, respectively. We illustrate our overall framework in Figure \ref{fig:framework}.

\begin{figure*}[t]
\begin{center}
\includegraphics [width=1\textwidth]{fig/fig4_pami.pdf}
\caption{The proposed CD-GAME framework consists of three modules: CD-ATTACK (red), CD-DEFENSE (green), and the Surrogate Community Detection Model (black). The attacker utilizes a learnable attack mask to map the clean graph $G$ to an attack graph $\hat{G}$, which is updated based on the gradient from the attack objective $\mathcal{L}_a$.  The defender uses a defense mask to generate a defense graph $\bar{G}$ guided by the defense objective $\mathcal{L}_d$. The defense graph is fed into the surrogate community detection model to obtain community detection results.   In CD-GAME, the attacker also considers its indirect influence (blue dashed line) on the defender, as the defender observes the attacker's operations when deciding their defense plan.}
\label{fig:framework}
\end{center}
\end{figure*}

\subsection{Surrogate Community Detection Model}\label{SCDM}
We propose a novel graph learning-based community detection model.  There are two key issues: (1) a loss function to measure the quality of the result, and (2) a neural network to detect communities.  We discuss these two issues below.

\subsubsection{The loss function}\label{dem}

We generalize \emph{normalized cut} \cite{shi2000normalized}, a widely used community detection measure, to serve as the loss function of neural networks.  Normalized cut measures the volume of edges being removed due to graph partitioning:
\begin{equation}
  \frac{1}{K}\sum_{k}\frac{cut(V_k,\overline{V_k})}{vol(V_k)},
\label{equ.H1}
\end{equation}
where $vol(V_k) = \sum_{i \in V_k}degree(i)$, $\overline{V_k} = V\setminus V_k$, $cut(V_k,\overline{V_k}) = \sum_{i \in V_k, j \in \overline{V_k}} A_{ij}$.  The numerator counts the number of edges between a community $V_k$ and the rest of the graph, while the denominator counts the number of incident edges to nodes in $V_k$.  Let $C \in \mathbb{R}^{N\times K}$ be a community assignment matrix where $C_{ik} = 1$ represents node $i$ belonging to the community $k$ and $0$ otherwise.  As we focus on detecting non-overlapping communities, an explicit constraint is that $C^\top C$ is a diagonal matrix.  With $C$,  normalized cut can be re-written as:
\begin{equation}
\begin{split}
 & \frac{1}{K}\sum_{k} \frac{\sum_{k^{\prime} \neq k} C_{:, k}^T A C_{:, k^{\prime}}}{C_{:, k}^T D C_{:, k}}\\
= &\frac{1}{K}\sum_{k}\frac{C_{:,k}^\top A(\boldsymbol{1}-C_{:,k})}{C_{:,k}^\top DC_{:,k}},
\end{split}
\label{equ.H2}
\end{equation}
where $D$ is the degree matrix with $D_{ii} = \sum_j A_{ij}$, $C_{:,k}$ is the $k$-th column vector of $C$.  We get the following simplified target function:
\begin{equation}
\begin{split}
  &\frac{1}{K}\sum_{k}\frac{C_{:,k}^\top LC_{:,k}}{C_{:,k}^\top DC_{:,k}}\\
  =&\frac{1}{K}\operatorname{Tr}((C^\top LC)\oslash(C^\top DC)),
\end{split}
\label{equ.H3}
\end{equation}
where $L = D -A$ is the Laplacian matrix, $\oslash$ is element-wise division, $\operatorname{Tr}(\cdot)$ is defined as the sum of elements on the main diagonal of a given square matrix.

Note that $C^\top C$ needs to be a diagonal matrix, we hereby introduce a new penalization term which explicitly incorporates the constraint on $C$ into the target function.  We subtract $C^\top C$ by $I_K$ as a penalization, where $I_K$ is an identity matrix.  Thus we define the differentiable unsupervised loss function $\mathcal{L}_u$ as follows:
\begin{equation}
  \mathcal{L}_u = \frac{1}{K}\operatorname{Tr}((C^\top LC)\oslash(C^\top DC)) + \gamma\big|\big|\ \frac{K}{N}C^\top C - I_K\ \big|\big|_F^2,
\label{equ.Hu}
\end{equation}
where $\big|\big|\cdot\big|\big|_F$ represents the Frobenius norm of a matrix, $\frac{K}{N}$ is applied as normalized cut encourages balanced clustering and voids shrinking bias \cite{shi2000normalized,tang2018normalized}. To analyze why the proposed penalization is valid, we assume that there are two communities $i, j$ to be clustered with two assignment vectors $C_{:,i}, C_{:,j}$ respectively.  Suppose each row of $C$ is a distribution, and for a non-diagonal element $(C^\top C)_{ij} = \sum_xC_{x, i}C_{x, j}$, it has a maximum value when nodes are assigned uniformly and a minimum value when nodes are assigned discriminatively.  By minimizing this penalization term, we actually encourage nodes to be discriminatively assigned to different communities. 

\subsubsection{The network architecture}\label{ssc}
This part details the network architecture to derive the community assignment matrix $C$, which is trained by minimizing the unsupervised loss $\mathcal{L}_u$ in Eq.\ \ref{equ.Hu}. In the literature, spectral clustering \cite{shi2000normalized} is usually used to compute the community assignment matrix $C$.  Due to its limitations in scalability and generalization, some recent work \cite{shaham2018spectralnet} has used a deep learning approach to overcome these shortcomings.  Our architecture, which is based on the recent popular graph neural networks (GNNs) \cite{kipf2017semi,klicpera2018predict} on graph data, has two main parts: (1) node embedding and (2) community assignment. In the first part, we leverage GNNs to get topology-aware node representations, so that similar nodes have similar representations. In the second part, based on the node representations, we assign similar nodes to the same community.    
Below, we introduce each component.

In this work, we use graph convolutional networks (GCNs) \cite{kipf2017semi} for the purpose of embedding nodes.  In the preprocessing step, the adjacency matrix $A$ is normalized:
\begin{equation}
  \bar{A} = \tilde{D}^{-\frac{1}{2}}(A + I_N)\tilde{D}^{-\frac{1}{2}},
\label{equ.gene}
\end{equation}
where $I_N$ is the identity matrix and $\tilde{D}_{ii} = \sum_j\ (A + I_N)_{ij}$. Then we transform node features over the graph structure via two-hop smoothing:
\begin{equation}
  H_l = \bar{A} \sigma (\bar{A}XW^0)W^1,
\label{equ.H}
\end{equation}
where $\sigma(\cdot)$ is the activation function such as $ReLU(\cdot)$, $W^0 \in \mathbb{R}^{d \times h}$ and $W^1 \in \mathbb{R}^{h \times v}$ are two weight matrices.  Intuitively, this function can be considered as a Laplacian smoothing operator \cite{jiawww19} for the features of the nodes in graph structures, which makes the nodes more proximal if they are connected within two hops in the graph.  With the node representations $H_l$ in hand, we are now ready to assign similar nodes to the same community by:
\begin{equation}
  C = \textsf{softmax}( \sigma (H_lW_{c1})W_{c2}),
\label{equ.C}
\end{equation}
where $W_{c1} \in \mathbb{R}^{v \times r}$ and $W_{c2} \in \mathbb{R}^{r \times K}$ are two weight matrices.  The function of $W_{c1}$ is to linearly transform the node representations from a $v$-dimensional space to a $r$-dimensional space, then nonlinearity is introduced by tying with the function $\sigma$. $W_{c2}$ is used to assign a score to each of the $K$ communities. Then \textsf{softmax} is applied to derive a standardized distribution for each node over the $K$ communities, which means that the sum of the $K$ scores for each node is $1$.

In summary, we design a neural network to serve as the surrogate community detection model, which is trained in an unsupervised way based on the loss function $\mathcal{L}_u$.  

\subsection{CD-ATTACK}\label{CGG}
An adversarial graph should achieve the goal of potential community detection attack, while, at the same time, it should be as \emph{similar} to the original graph as possible to remain stealthy.  In the literature, several works measure the similarity between graphs by the number of discrete changes. For example, \cite{sun2018adversarial, dai2018adversarial} limit the number of allowed changes on edges by a budget $\epsilon$:
\begin{equation}
  \sum_{i<j}|A_{ij} - \hat{A}_{ij}| \leq \epsilon.
\label{equ.H7}
\end{equation}

In this subsection, we describe our method, which produces a modified graph $\hat{A}$ that meets the budget constraint in Eq.\ \ref{equ.H7}.  However, generating such an attacked graph is challenging due to the following issues:
\begin{itemize}
\item  \emph{Budget constraint}: Generating a valid graph structure with a budget constraint is difficult, as it is a complex combinatorial optimization problem.
\item  \emph{Scalability}: Existing graph generation methods suffer from scalability issues \cite{wu2019comprehensive}.
\end{itemize}

To solve the complex discrete combinatorial optimization problem with a \emph{budget constraint}, we reformulate it into a locally differentiable problem. We propose to directly incorporate prior knowledge about the graph structure using a \emph{Learnable Editing Mask} and conduct efficient goal-oriented malicious edge editing. Furthermore, our method shows promising performance in \emph{scalability}, as the computation focuses on the scope of edge perturbation rather than the entire graph structure.

\subsubsection{Learnable Editing Mask}\label{sec:lem}

To facilitate a gradient-based approach that iteratively adjusts the intermediate attack strategy under the guidance of a malicious purpose, we formulate a \emph{Learnable Editing Mask} $M \in \{\sigma(m) \ | \ m \in \mathbb{R}\}^{|E_u|}$, where $\sigma(\cdot)$ is the Sigmoid function and $E_u$ is the attack scope of the target node $u$.  In other words, $E_u$ defines the operation space of the target node $u$ for adding and removing edges.  Each element in $M$ indicates the priority of the corresponding candidate edge perturbation in $E_u$. Given budget constraints $\epsilon$, the Learnable Editing Mask $M$ can be mapped to a discrete attack strategy $\tilde{M}  \in \{0,1\}^{|E_u|}$ corresponding to the top-$\epsilon$ elements in $M$ or by sampling a $\tilde{M}$ with respect to the weights in $M$.

Another challenge is to determine the type of edge perturbation (adding or deleting an edge). To this end, we introduce the \emph{Operation Indicator Vector} $N$  to indicate which perturbation type is conducted on the graph structure. For those edges that already exist in the original social network, we denote them as $E_{edge} \subseteq  E_u$. Similarly, we use $E_{nonedge} \subseteq  E_u$ to represent empty edges that did not exist before.   Then, the Operation Indicator Vector $N$ is represented as $N = [N_{nonedge}, N_{edge}] $, where $N_{nonedge}$ is $\{1\}^{|E_{nonedge}|}$ and $N_{edge}$ is $\{-1\}^{|E_{edge}|}$. This Operation Indicator Vector $N$ guarantees that the edge perturbation is valid. Specifically, we only allow the removal of existing edges and the addition of non-existing edges.

Finally, by integrating the discrete attack strategy $\tilde{M}$ with the operation indicator vector $N$, we can obtain the perturbed structure $\hat{E}_u$:
\begin{equation}
  \hat{E}_u = E_u + \tilde{M} \otimes  N,
\end{equation}
where $\otimes$ denotes the element-wise product. Consequently, we can derive the final perturbed global graph structure $\hat{A}$ based on $\hat{E}_u^k$.

\subsubsection{Goal-Oriented Malicious Attack}
Here, we introduce our attack objective function $\mathcal{L}_a$ to diverge the community assignment distributions within the set of individuals:
\begin{equation}
  \mathcal{L}_{a} = - \min_{i \in C^+,j \in C^+} KL(C_{i,:}||C_{j,:}),
\label{eau:attobj}
\end{equation}
where $\mathcal{L}_{a}$ can be regarded as the margin loss or the smallest distance for any pair inside $C^+$, and we aim to maximize the margin so that the members of $C^+$ are spread out across the communities. Accordingly, the attacker can adjust the attack mask $M_a$ following the basic update rule as follows:
\begin{equation}
  M_a \gets M_a - \eta^a \operatorname{sign}(
  \frac{\partial \mathcal{L}_a}{\partial \tilde{M}_a}),
\end{equation}
where $\eta^a$ is the step size and $\operatorname{sign}(\cdot)$ is the sign 
 function. In summary, the learnable editing mask receives an error signal $\mathcal{L}_{a}$ from the surrogate community detection model, which serves as a goal to guide the generation of attack graphs.

\subsection{CD-DEFENSE}

In this section, our goal is to build a robust community detection system that can withstand possible imperceptible structural perturbations. In general, nodes share common properties and/or play similar roles within the social network~\cite{fortunato2010community}. For instance,  social media users who have similar interests often create their own discussion groups. 
To mitigate the effects of adversarial perturbations, we adopt a strategy that is the opposite of the attack. 
More specifically,
we tend to \emph{add connections between nodes within the same communities and delete edges across different communities}. 
However, community detection is an unsupervised problem, making it difficult to determine the ground truth of community assignment. Therefore, a question arises: \emph{How can we implement the defense strategy without the supervision?}

It has been observed that malicious attacks tend to destroy this similarity by adding or deleting connections between nodes from different or the same communities, which provides the most benefit to the adversarial objective~\cite{wu2019adversarial}.  In graph spectrum energy distribution, adversarial perturbations transfer low-frequency energy to the high-frequency part \cite{sandryhaila2013discrete}. The ‘right-shift’ of spectral energy has also been validated in graph anomaly detection as the fraction of anomalies increases \cite{tang2022rethinking}.
Therefore, we aim to reduce the high-frequency energy caused by adversarial attacks and ensure energy concentration in the low-frequency area. The \emph{Rayleigh Quotient} represents the high-frequency area of the social network \cite{tang2022rethinking}. Therefore, we minimize the \emph{Rayleigh Quotient} of the perturbed graph $\hat{G}$ to reduce its high-frequency energy:
\begin{equation}
\mathcal{L}_d = \frac{\frac{1}{2} \sum_{i=1}^N \sum_{j=1}^N \hat{A}_{i j}\left(x_i-x_j\right)^2}{\left \|x\right \|^2_2}   = \frac{x^\top \hat{L} x}{x^\top x},
\label{RayleighQuotient}
\end{equation}
where $x$ is the signal on the social network and $\hat{L} = \hat{D}-\hat{A}$ is the Laplacian matrix of the perturbed graph. 

\textbf{Remark.}  We provide an intuitive understanding of the role of Eq. \ref{RayleighQuotient} for the two actions: (1) Edge addition. By adding an edge, the numerator of the Rayleigh Quotient is augmented by a term 
$\left(x_i-x_j\right)^2$, thus minimizing Eq. \ref{RayleighQuotient} is equivalent to finding the minimum $\left(x_i-x_j\right)^2$ that satisfies $A_{ij}=0$. Taking into account the fact that communities are groups of vertices similar to each other, it tends to add connections between nodes within the same communities. (2) Edge deletion. Similarly, deleting an edge is equivalent to finding the maximum  $\left(x_i-x_j\right)^2$ that satisfies $A_{ij}=1$. It tends to remove connections between nodes within the different communities.

To build a robust community detection system, we use 
another learnable defense editing mask $M_d$ and its discrete version $\Tilde{M}_d$ as described in Sec.~\ref{sec:lem}.  Accordingly, the defender can adjust the learnable editing mask $M_d$ following the basic update rule as follows:
\begin{equation}
  M_d \gets M_d - \eta^{d} \operatorname{sign}(\frac{\partial \mathcal{L}_d}{\partial \tilde{M}_d}),
\end{equation}
where $\eta^{d}$ is the step size. Therefore, our defender is capable of adaptively modifying the network structure to reduce the high-frequency energy increased by adversarial attacks.

\subsection{ CD-GAME}
In general, attack and defense behaviors usually manifest as an interactive process, where the attacker initiates a strategy, followed by the defender reinforcing their detection algorithms.  In this section, we define this interactive attack-defense process as a \emph{Stackelberg game}. We set the attacker as the ‘leader’ and the defender as the ‘follower’. The designation of ‘leader’ and ‘follower’ indicates the order of play between the players, meaning the leader plays first and the follower second. For brevity, we refer to $\tilde{M}$ as $M$ in this section. In this game, the attacker and defender aim to solve the following problems:
\begin{equation}\label{sg}
\begin{aligned}
& \text{(A)} \ \ \min _{M_a} \ \{\mathcal{L}_a\left(M_a, M_d\right) \mid M_d \in \arg \min _{y} \mathcal{L}_d\left(M_a, y \right)\}, \\
& \text{(D)} \ \ \min _{M_d} \ \mathcal{L}_d\left(M_a, M_d \right). 
\end{aligned}
\end{equation}
This is different from the \emph{simultaneous} play game in which each player is faced with an optimization problem of the same form as the defender in Eq.~\ref{sg} (D). In contrast, the attacker accounts for the possible actions of the defender, involving mutual influence and feedback mechanisms between players. 

To solve this problem and take into account the mutual influence between the attacker and defender, we define an implicit map $r: M_a \mapsto M_d$. Therefore, we can rewrite $\mathcal{L}_a\left(M_a, M_d\right)$ as $\mathcal{L}_a\left(M_a, r(M_a)\right)$.  Since it is conditioned on $M_d \in \arg \min _{y} \mathcal{L}_d\left(M_a, y \right)$, we have $\frac{\partial \mathcal{L}_d}{ \partial M_d } = 0$. Naturally, we could obtain the total derivative of $\mathcal{L}_a$ via invoking the implicit function theorem: 
\begin{equation}
\frac{\mathrm{d} \mathcal{L}_a}{\mathrm{d} M_a}  = \frac{\partial \mathcal{L}_a}{\partial M_a}-\frac{\partial^2 \mathcal{L}_d ^\top}{\partial M_d  \partial M_a}  \frac{\partial^2 \mathcal{L}_d ^{-1}}{ \partial M_d^2} \frac{\partial \mathcal{L}_a}{\partial M_a}.
\end{equation}
Accordingly, we solve the Stackelberg game following the rule below:
\begin{equation}
\begin{aligned}
  &M_a \gets M_a - \eta^a \operatorname{sign}(
 \frac{\mathrm{d} \mathcal{L}_a}{\mathrm{d} M_a}), \\
   &M_d \gets M_d - \eta^d \operatorname{sign}(\frac{\partial \mathcal{L}_d}{\partial M_d}).
\end{aligned}
\label{gameupdate}
\end{equation}
Our proposed CD-GAME is illustrated in Alg. \ref{algo.game}. We alternatively update the attack mask and defense mask until they achieve convergence.

\begin{algorithm}
  \caption{CD-GAME}
  \label{algo.game}
  \KwIn{Original graph $G$, perturbation budget $\epsilon$ and target nodes $C^+$.}
  \KwOut{Surrogate community detection model $\theta$, attack graph $\hat{G}$ and defense graph $\bar{G}$.}
  Initial: parameters $M_a, M_d$;
  
  // Pre-training a community detection model;
  
  \Repeat{No. of epochs}{
    
    $G \leftarrow$ full batch;
    
	Compute $\mathcal{L}_{u}$;

    $\theta \leftarrow \theta - \eta^{\theta} \frac{\partial \mathcal{L}_{u}}{\partial \theta}$\;
  }
 
  // Security game;
  
  \Repeat{No. of epochs}{
    
    $G \leftarrow$ full batch\;
    $\tilde{M}_a,\tilde{M}_d \leftarrow \operatorname{discretization}(\{M_a,M_d\}) $\;
	Attack graph $\hat{G} \leftarrow \operatorname{Edit}(\tilde{M}_a, G )$\;
 Defense graph $\bar{G} \leftarrow \operatorname{Edit}(\tilde{M}_d, \hat{G} )$\;
 Compute $\mathcal{L}_a$ and $\mathcal{L}_d$ with $\{\hat{G},\bar{G}\}$\;

// Update parameters;
    
    $M_a \gets M_a -  \eta^a \operatorname{sign}(
 \frac{\partial \mathcal{L}_a}{\partial \tilde{M}_a}-\frac{\partial^2 \mathcal{L}_d ^\top}{\partial \tilde{M}_d  \partial \tilde{M}_a}  \frac{\partial^2 \mathcal{L}_d ^{-1}}{ \partial \tilde{M}_d^2} \frac{\partial \mathcal{L}_a}{\partial \tilde{M}_a})$\;
    $M_d \gets M_d - \eta^d \operatorname{sign}(\frac{\partial \mathcal{L}_d}{\partial \tilde{M}_d})$\; }

\end{algorithm}

In Stackelberg game, a crucial concept lies in characterizing the existence of the \emph{Stackelberg equilibrium}. The Stackelberg equilibrium is the point at which both the leader and the follower have made their decisions, and neither of them can benefit from changing their decision, given what the other player has decided. A natural question is whether a Stackelberg equilibrium exists in CD-GAME. To address this, we present the following theoretical result.

\begin{definition}[Stackelberg Equilibrium]
Consider $M_a \in U_a$ for attacker (leader) and  $M_d \in U_d$ for  defender (follower). The strategy $M_a^*$ is a Stackelberg solution for the leader if, $\forall M_a \in U_a$,
$$\sup _{M_d \in R_{U_d} \left(M_a^*\right)} \mathcal{L}_a \left(M_a^*, M_d\right) \leq \sup _{M_d \in R_{U_d} \left(M_a\right)} \mathcal{L}_a\left(M_a, M_d\right),$$
where $R_{U_d} \left(M_a\right)=\left\{y  \mid \mathcal{L}_d \left(M_a, y\right) \leq \mathcal{L}_d \left(M_a, M_d\right), \forall M_d \in\right.$ $\left.U_d\right\}$. Moreover, $\left(M_a^*, M_d^*\right)$ for any $M_d^* \in R_{U_d}\left(M_a^*\right)$ is a Stackelberg equilibrium on $U_a \times U_d$.
\end{definition}

\begin{theorem}
A Stackelberg equilibrium exists between the leader (attacker) and follower (defender) in CD-GAME.
\end{theorem}
\begin{proof}
Let us consider any $M_a \in U_a$. Given that $U_d$ is defined on the perturbation scope, the cardinality of $U_d$ is finite (i.e., $|U_d|< \infty$). There exists at least one $M'_d$ such that $M'_d = \arg \min \mathcal{L}_a\left(M_a, M'_d\right)$. Consequently, every $M'_d$ together forms a non-empty finite set $R_{U_d} \left(M_a\right)$ that satisfies the following condition, $\forall M'_d \in R_{U_d} \left(M_a\right)$,
$$
\mathcal{L}_d \left(M_a, M'_d\right) \leq \mathcal{L}_d \left(M_a, M_d\right).
$$
We can define the set $ R_{U_d}$ for all $M_a \in U_a$ as $ R_{U_d} = \left\{ R_{U_d}\left(M_a\right)  \mid M_a \in U_a \right \}$. Given that the cardinality of $U_a$ is finite (i.e., $|U_a|< \infty$), it follows that $| R_{U_d}|< \infty$. Therefore, there exists a $M_a^*$ that satisfies:
$$
 M_a^* = \mathop{\arg\min}_{M_a} \sup _{M_d \in R_{U_d} \left(M_a\right)} \mathcal{L}_a \left(M_a, M_d\right). 
$$
Hence, the strategy $M_a^*$ is a Stackelberg solution for the leader (attacker). Given that $R_{U_d} \left(M_a^*\right)$ is non-empty, there exists a $M_d^* \in R_{U_d} \left(M_a^*\right)$ for the follower (defender). Therefore, a Stackelberg equilibrium exists in the CD-GAME and is given by the pair $\left(M_a^*, M_d^*\right)$.
\end{proof}

\section{EXPERIMENTS}\label{sec.exp}
In this section, we present our empirical findings across different graph datasets. In addition, we follow \emph{normalized cut} \cite{shi2000normalized} and evaluate our method on image segmentation.

\subsection{Data}
We validate the effectiveness of our model on six real-world graph data sets: (1) DBLP, (2) Cora, (3) Photos, (4) Citeseer, (5) Flickr, and (6) OGBN-Arxiv.  Table \ref{tab:twodatasets} lists the statistics of the six graph datasets.

\textbf{DBLP.} From DBLP data, we build a coauthor graph with 5,304 authors.  Its edges represent coauthor relationships.

\textbf{Cora.} The Cora dataset \cite{sen2008collective} is a citation network of machine-learning papers, organized into 7 distinct classes. 

\textbf{Photos.} The Photos dataset \cite{mcauley2015image} is from the Amazon co-purchase graph, where nodes represent goods, and edges indicate that two goods are frequently bought together. 

\textbf{Citeseer.}
The Citeseer dataset \cite{sen2008collective}  consists of scientific publications classified into 6 classes. 

\textbf{Flickr.}
The Flickr dataset originates from NUS-wide~\cite{zenggraphsaint} and consists of images classified into 7 classes. Edges indicate that two images share common properties. 

\textbf{OBGN-Arxiv.}
The OBGN-Arxiv dataset \cite{hu2020open} is a citation network among all Computer Science papers in arXiv, classified into 40 subject areas. 
\subsection{Metrics}\label{syn.base}

We follow \cite{waniek2018hiding} and use two measures to quantify the degree of hiding of target nodes:
\begin{equation}
M1(C^+,G) = \frac{|{G_i:G_i\cap C^+ \neq \emptyset}| - 1}{(K -1)\times \max_{G_i}(|G_i\cap C^+|)},
\end{equation}
where $K > 1$. The numerator grows linearly with the number of communities that $C^+$ is distributed over.  The denominator penalizes the cases in which $C^+$ is skewly distributed over the community structures. Intuitively, this measure $M1(C^+,G)$ focuses on how well the members of $C^+$ spread out across $G_1, \ldots, G_K$, and $M1(C^+,G) \in [0,1]$.
\begin{equation}
M2(C^+,G) = \sum_{G_i:G_i\cap C^+ \neq \emptyset}\frac{|G_i \setminus C^+|}{\max(N-|C^+|,1)},
\end{equation}
where the numerator grows linearly with the number of non-members of $C^+$ that appear simultaneously with members of $C^+$ in the same community. Note the numerator only counts the communities in which there exists a member of $C^+$. Basically, this measure focuses on how well $C^+$ is hidden in the crowd and $M2(C^+,G) \in [0,1]$.

For both measures, a greater value denotes a better attack performance.  Conversely, a smaller value denotes better defense performance.

\begin{table}[t]
\centering
  \caption{Statistics of datasets}
  \label{tab:twodatasets}
  \scalebox{0.9}{
  \begin{tabular}{ccccc}
    \hline
& No. of nodes&No. of edges&No. of features \\
    \hline
	\textbf{DBLP}&5,304&28,464&305\\
	\textbf{Cora} &2,708&10,556&1,433\\
	\textbf{Photos}&7,650&238,163&745\\
	\textbf{Citeseer} &3,327& 4,732& 3,703\\
    \textbf{Flickr} &89,250&  899,756& 500\\
    \textbf{OBGN-Arxiv} &169,343&  1,166,243& 128\\
  \hline
\end{tabular}}
\end{table}

\subsubsection{Setup}
We select the target nodes by the following process: (1) For DBLP without ground truth, we use Infomap \cite{rosvall2008maps} to divide it into 10 communities, and (2) in each community, we select 5 nodes with the highest degrees and another $5$ nodes randomly. The budget $\epsilon$ is equal to the number of target nodes unless specified. Our implementation is based on PyTorch.  We use an Adam optimizer with exponential decay. We set $\gamma = 0.1$ in Eq. \ref{equ.Hu}. The output dimension of the first-layer GCN is 32, and that of the second-layer GCN is 16.  We set the fully connected layer with 32 rectified linear units and a dropout rate of 0.3. The initial learning rate is 0.001. We average all the results across 5 independent runs.

\subsection{Malicious Attack}
In this section, we evaluate the effectiveness of CD-ATTACK by attacking the surrogate community detection model.  
\subsubsection{Baselines}
We use the following approaches as our attack baselines:
\begin{itemize}

\item \textbf{DICE} \cite{waniek2018hiding}. It randomly deletes edges incident to the target set $C^+$, then spends the remaining budget on inserting edges between $C^+$ and the rest of the graph.

\item \textbf{MBA} \cite{fionda2017community}. It randomly deletes intra-community edges and inserts inter-community edges. 

\item \textbf{RTA} \cite{zugner2018adversarial}. In each step, a node is randomly sampled. If connected to 
$C^+$, an edge between them is deleted; otherwise, an edge is added.

\item \textbf{CD-ATTACK-RL \cite{li2020adversarial}}. It is the reinforcement learning variant of CD-ATTACK, with the reward from Eq.~\ref{eau:attobj}.

\item \textbf{ProHiCo \cite{ProHiCo}}. It first allocates perturbation randomly, then greedily selects edges by likelihood minimization.

\item \textbf{DRL \cite{DRL}}. It frames the problem as a Markov decision process and solves it with deep reinforcement learning.

\end{itemize}

In the following section, \textbf{Clean} refers to the community detection results of the original clean graph.

\begin{table*}[t]
  \caption{Performance comparison of different attacks on the surrogate model}
  \centering
  \begin{tabular}{ccccccccccccc}
    \hline
Datasets&\multicolumn{2}{c}{\textbf{DBLP}}&\multicolumn{2}{c}{\textbf{Cora}}&\multicolumn{2}{c}{\textbf{Photos}}&\multicolumn{2}{c}{\textbf{Citeseer}}&\multicolumn{2}{c}{\textbf{Flickr}}&\multicolumn{2}{c}{\textbf{OGBN-Arxiv}}\\
\hline
       -      &M1&M2&M1&M2&M1&M2&M1&M2&M1&M2&M1&M2\\
       \hline
       \textbf{Clean} &3.61\%&34.11\%&3.25\%&53.55\%&6.91\%&50.77\%&5.37\%&58.32\% &4.63\%&47.35\% &3.88\%&43.97\%\\
    \hline
	\textbf{DICE} &4.23\%&36.22\%&3.25\%&53.55\%&6.91\%&50.77\%&5.21\%&59.61\% &5.08\%&48.16\% &4.01\%&45.44\%\\
	\textbf{MBA} &4.08\%&37.23\%&3.31\%&52.00\%&7.04\%&53.67\%&5.08\%&57.11\% &4.94\%&49.32\% &4.08\%&44.77\%\\
	\textbf{RTA} &4.45\%&35.45\%&3.08\%&52.66\%&7.15\%&52.30\%&5.33\%&56.72\% &5.27\%&46.74\% &3.96\%&46.42\%\\
    \textbf{ProHiCo} &4.98\% &38.95\% &3.12\% &55.79\% &7.17\% &55.92\% &5.36\% &61.14\% &5.91\% &49.88\% &4.37\% &48.90\% \\
    \textbf{DRL} &5.31\% &39.28\% &3.45\% &60.12\% &7.50\% &56.35\% &5.69\% &61.47\% &6.24\% &50.21\% &4.40\% &49.33\% \\
	\hline
    \textbf{CD-ATTACK-RL} &5.02\%  &39.73\% &3.31\% &59.69\% &7.73\% &55.98\% &5.75\% &61.03\% &5.82\% &49.77\% &4.35\% &48.98\%\\
	\textbf{CD-ATTACK} &\textbf{6.06\%}&\textbf{44.89\%}&\textbf{3.82\%}&\textbf{64.52\%}&\textbf{8.11\%}&\textbf{57.62\%}&\textbf{6.33\%}&\textbf{63.72\%}&\textbf{7.18\%}&\textbf{54.16\%}&\textbf{5.73\%}&\textbf{56.56\%}\\
  \hline
\end{tabular}
 \label{table: attack performance}
\end{table*}

\begin{figure*}[!t]
    \centering
    \subfloat[Attack M1\label{fig:budget1}]{\includegraphics[width=0.25\textwidth]{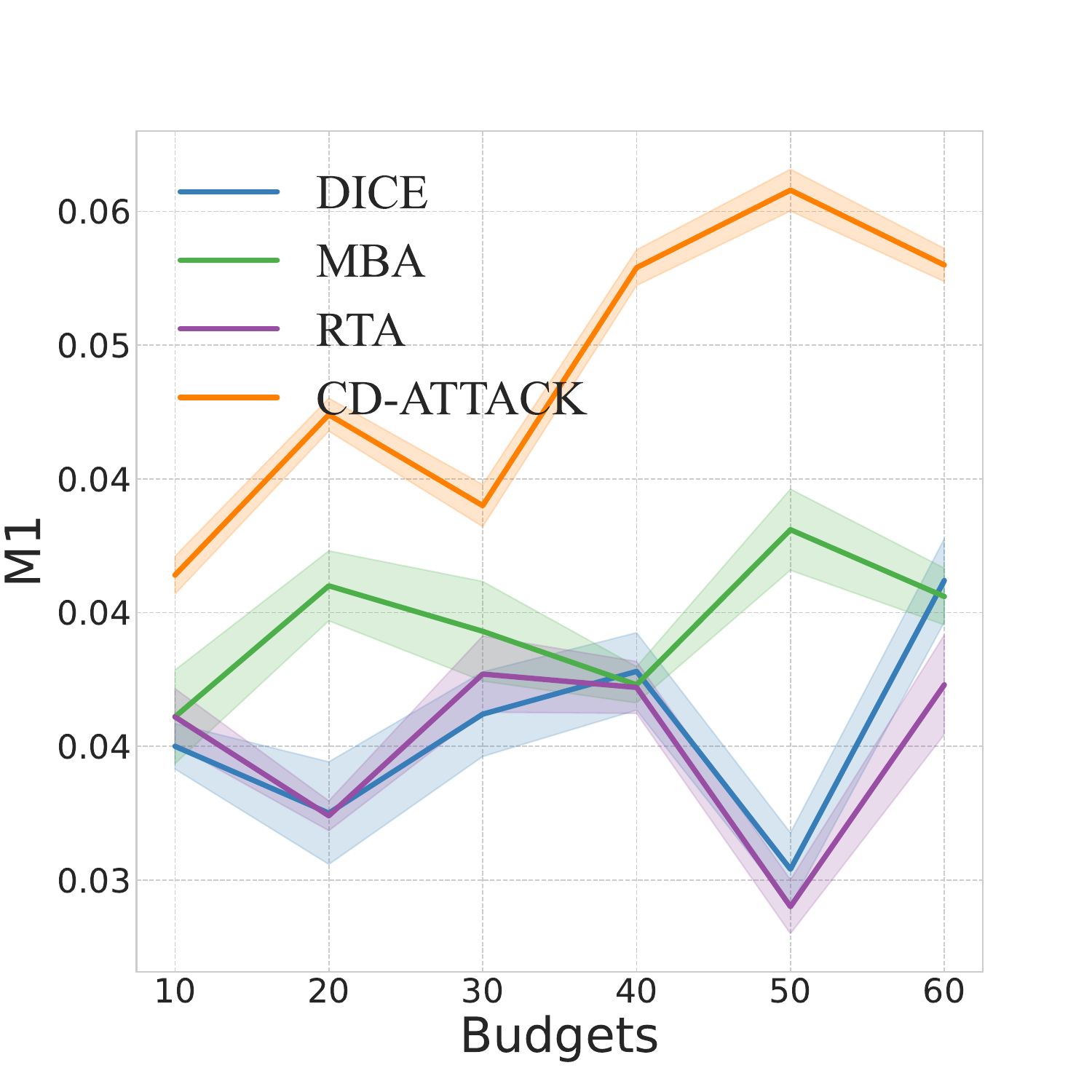}}  
    \subfloat[Attack M2\label{fig:budget2}]{\includegraphics[width=0.25\textwidth]{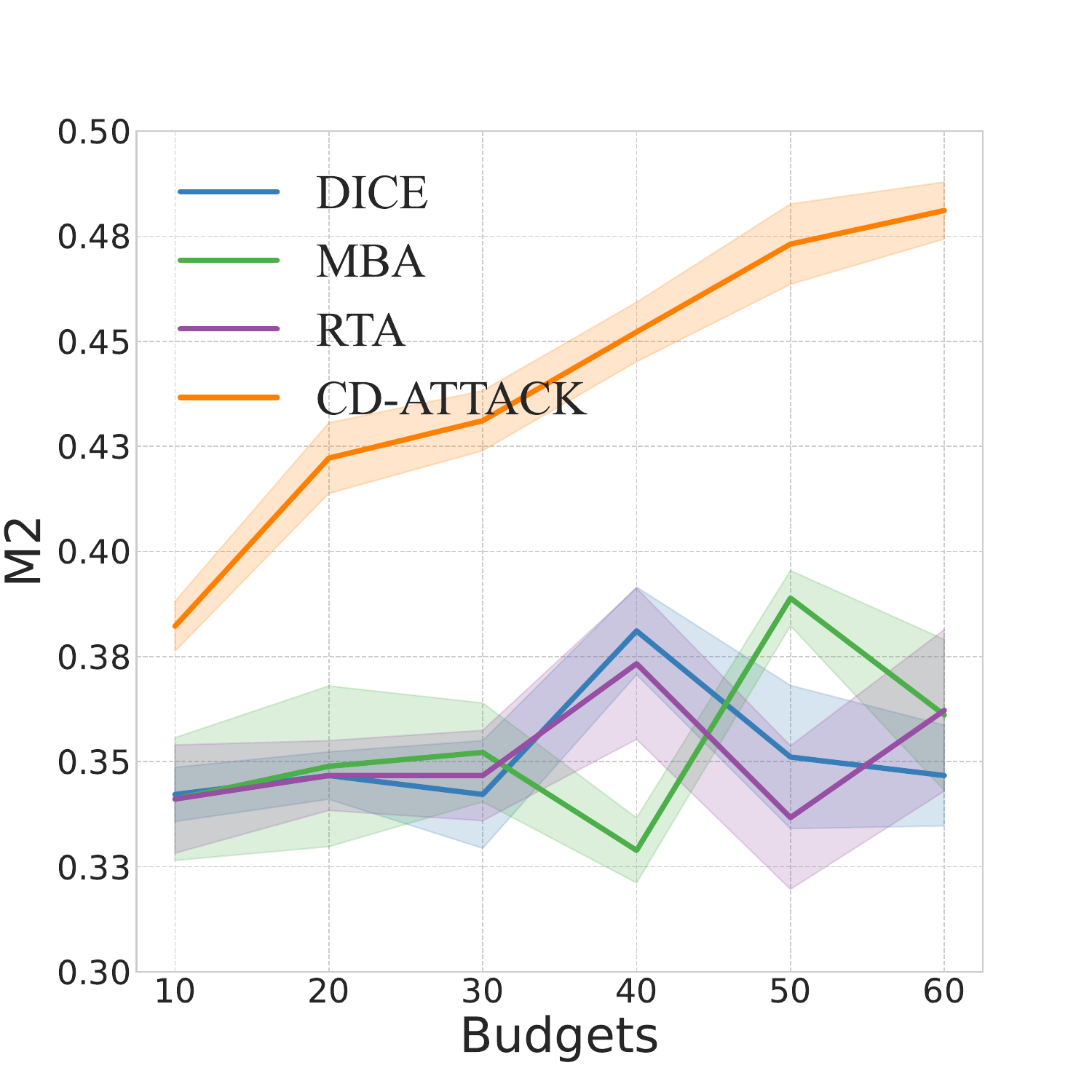}}
    \subfloat[Defense M1\label{fig:budget3}]{\includegraphics[width=0.25\textwidth]{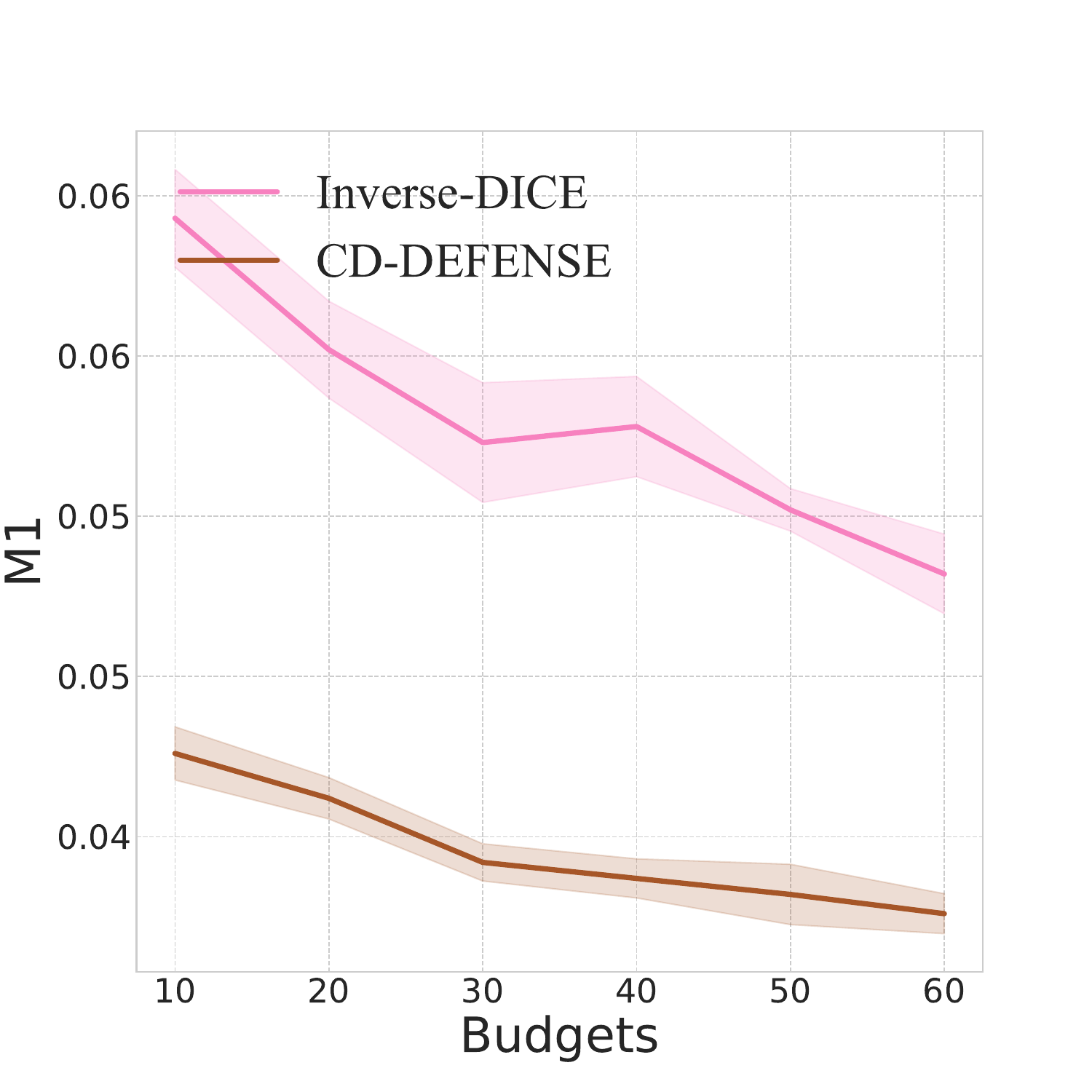}}
    \subfloat[Defense M2\label{fig:budget4}]{\includegraphics[width=0.25\textwidth]{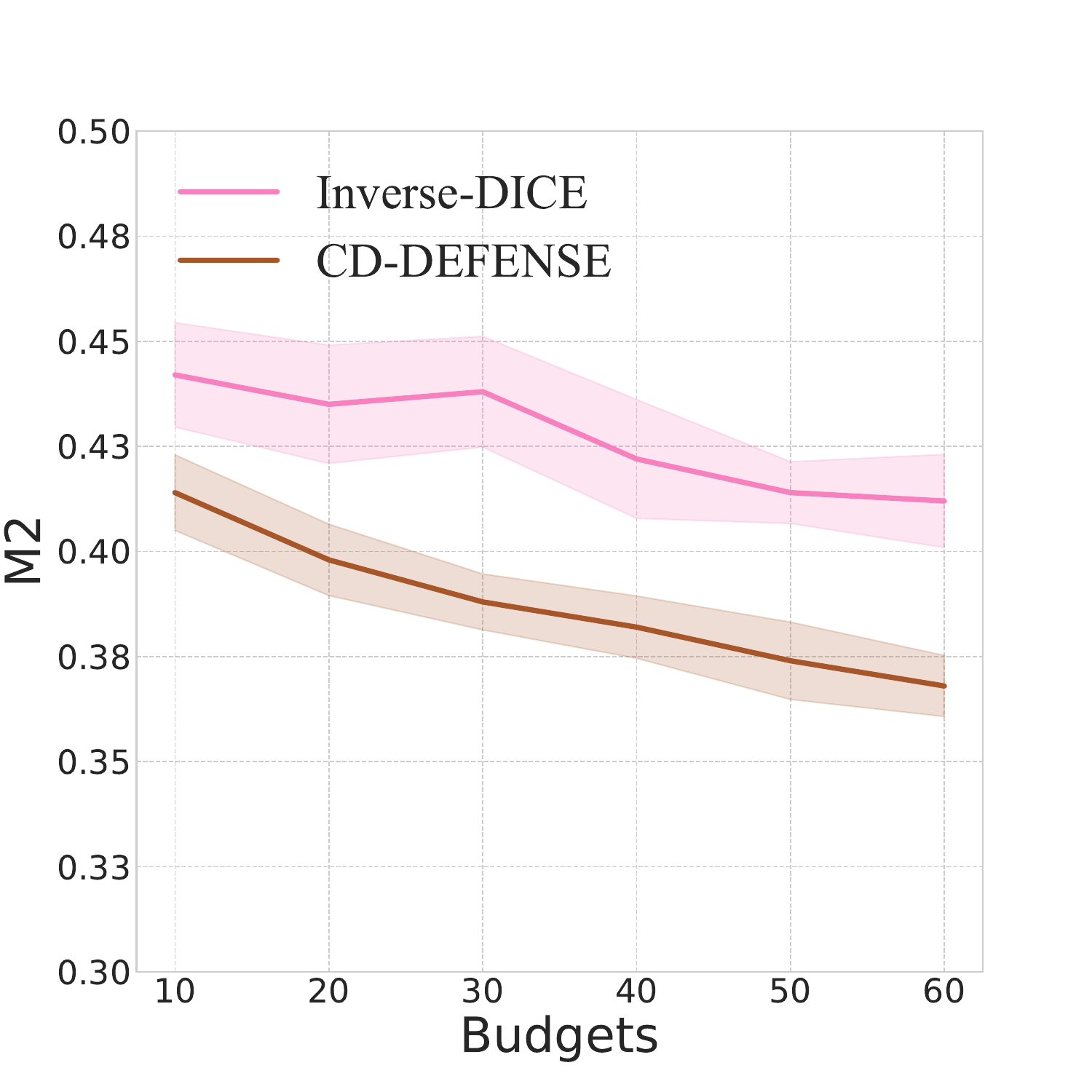}}
    \caption{Attack and Defense performance with different budgets on \emph{DBLP}.}
    \label{fig:budget}
\end{figure*}

\subsubsection{Attack performance}\label{sec: attack performance}

In Table \ref{table: attack performance}, CD-ATTACK achieves the best performance on all data sets in terms of both measures. DICE, MBA, and RTA are heuristic methods that depend on specific rules, leading to a coarse identification of relevant edges. ProHiCo employs a greedy strategy, which often gets trapped in a suboptimal solution. DRL and CD-ATTACK-RL approximate the problem as a Markov Decision Process (MDP) and apply reinforcement learning. However, they struggle with the vast search space. Therefore, whether due to coarse heuristics, greedy choices, or MDP approximations, these methods are more likely to overlook critical edges under a limited budget. In contrast, CD-ATTACK treat all egde perturbations as a whole, leverages the gradient to precisely identify the most impactful edges.

\subsubsection{Computational Complexity}

\begin{table}[h]
  \caption{Computational complexity}
  \label{tab:is}
  \centering
  \begin{tabular}{c|ccc}
    \hline
    Datasets &\textbf{DBLP}&\textbf{Citeseer}&\textbf{Flickr}\\
    \hline
    - &\multicolumn{3}{c}{Wall clock time (hour)} \\
    \hline
    \textbf{CD-ATTACK-RL} & 4.74 &  0.89 & 83.61\\
    \textbf{CD-ATTACK} & 0.03 &   0.01&  1.43\\
    \hline
\end{tabular}
 \label{table.Computationalcomplexity}
\end{table}

CD-ATTACK demonstrates significant advantages over CD-ATTACK-RL in computational efficiency. In Table \ref{table.Computationalcomplexity}, we list the wall clock time for CD-ATTACK and CD-ATTACK-RL to achieve convergence on NVIDIA V100 GPU. CD-ATTACK-RL is computationally expensive due to the large state-action space and the sample inefficiency common in reinforcement learning, requiring numerous interactions with the graph and surrogate model to learn an effective policy. In contrast, CD-ATTACK is efficient with gradient descent.

\subsubsection{Effect of Attack Budget}

We evaluate how the budget $\epsilon$ affects attack performance. Taking attacks on DBLP as an example, we vary $\epsilon$ from 10 to 60 and plot the corresponding attack performance in terms of M1 and M2 in Figures \ref{fig:budget1} and \ref{fig:budget2}. We find that our CD-ATTACK achieves the best attack performance across all budgets. As we increase $\epsilon$, the attack performance of CD-ATTACK improves (i.e., both M1 and M2 increase), indicating that a larger budget is beneficial for a successful attack.  Another observation is that the performance of the other baselines does not exhibit the same sensitivity to budget changes. This serves as additional evidence that the three baselines face challenges in identifying the critical edges that significantly increase attack effectiveness.

\subsubsection{Effect of Attack Scope}
We evaluate how the attack scope affects performance. We expand the attack scope to include 1-hop, 2-hop, and 3-hop neighborhoods and list the corresponding M1 and M2 in Table \ref{table:scope}. Our CD-ATTACK achieves the best performance across all attack scopes. As we expand the neighborhood range, all methods demonstrate a certain degree of improvement in effectiveness, suggesting that a broader attack scope encompasses more vital edges for successful attacks. Notably, the performance of CD-ATTACK significantly improves with the expansion of the neighborhood range.

\begin{table}[h]
  \caption{Effect of attack scope}
  \centering
  \resizebox{0.5\textwidth}{!}{
  \begin{tabular}{c|cccccc}
    \hline
Neighbor&\multicolumn{2}{c}{\textbf{N=1}}&\multicolumn{2}{c}{\textbf{N=2}}&\multicolumn{2}{c}{\textbf{N=3}}\\
\hline
       -      &M1&M2&M1&M2&M1&M2\\
       \hline
       \textbf{Clean} &3.61\%&34.11\%&3.61\%&34.11\%&3.61\%&34.11\%\\
       \hline
       	\textbf{DICE} &4.23\%&36.22\%&4.63\%&39.32\%&5.41\%&42.77\%\\
	\textbf{MBA} &4.08\%&37.23\%&5.02.\%&42.00\%&5.54\%&44.67\%\\
	\textbf{RTA} &4.45\%&35.45\%&4.73\%&42.66\%&5.37\%&43.35\%\\
    \textbf{ProHiCo}  &4.98\% &38.95\% &5.87\% &48.30\% &7.02\% &54.08\% \\
    \textbf{DRL} &5.31\% &39.28\% &6.33\% &52.25\% &7.38\% &56.15\% \\
         \hline
       \textbf{CD-ATTACK} &\textbf{6.06\%}&\textbf{44.89\%}&\textbf{7.21\%}&\textbf{58.64\%}&\textbf{8.36\%}&\textbf{63.42\%}\\
  \hline
\end{tabular}
}
 \label{table:scope}
\end{table}

\subsection{Defense}
In this section, we evaluate the effectiveness of CD-DEFENSE in mitigating the effects of adversarial attacks.

\begin{table*}[!h]
  \caption{Performance comparison of different defenses on the surrogate model}
  \centering
  \begin{tabular}{ccccccccccccc}
    \hline
Datasets&\multicolumn{2}{c}{\textbf{DBLP}}&\multicolumn{2}{c}{\textbf{Cora}}&\multicolumn{2}{c}{\textbf{Photos}}&\multicolumn{2}{c}{\textbf{Citeseer}}&\multicolumn{2}{c}{\textbf{Flickr}}&\multicolumn{2}{c}{\textbf{OGBN-Arxiv}}\\
\hline
       -      &M1&M2&M1&M2&M1&M2&M1&M2&M1&M2&M1&M2\\
       \hline
       \textbf{Clean} &3.61\%&34.11\%&3.25\%&53.55\%&6.91\%&50.77\%&5.37\%&58.32\% &4.63\%&47.35\% &3.88\%&43.97\%\\
       \textbf{CD-ATTACK} &6.06\%&44.89\%&3.82\%&64.52\%&8.11\%&57.62\%&6.33\%&63.72\% &7.18\%&54.16\%&5.73\%&56.56\%\\
    \hline
	\textbf{Inverse-DICE} &4.56\%&39.32\%&3.55\%&57.82\%&7.44\%&56.33\%&5.87\%&62.32\% &5.89\% &53.27\% &4.85\% &50.68\%\\
	\textbf{NE} &4.32\%&38.75\%&3.62\%&58.54\%&7.57\%&54.32\%&5.95\% &61.93\% &5.93\% &53.32\% &4.81\% &50.72\%\\
	\textbf{RS} &5.73\%&45.62\%&3.91\%&62.63\%&7.93\%&54.64\%&6.24\%&62.56\% &6.72\% &53.71\% &5.28\% &54.11\% \\
    \textbf{RobustECD-GA} &4.05\% &38.12\% &3.78\% &58.19\% &7.49\% &53.08\% &5.93\% &61.48\% &5.38\% &52.45\% &4.12\% &47.65\% \\
    \textbf{RobustECD-SE} &4.12\% &38.05\% &3.85\% &59.26\% &7.53\% &53.15\% &5.89\% &61.41\% &5.22\% &51.61\% &4.28\% &48.50\%\\
	\hline
	\textbf{CD-DEFENSE} &\textbf{3.58\%}&\textbf{35.67\%}&\textbf{3.32\%}&\textbf{55.73\%}&\textbf{7.02\%}&\textbf{51.62\%}&\textbf{5.47\%}&\textbf{60.02\%} &\textbf{4.87\%}&\textbf{50.17\%} &\textbf{3.94\%}&\textbf{47.06\%}\\
  \hline
\end{tabular}
 \label{table:defense}
\end{table*}

\begin{table*}[h]
  \caption{Performance comparison of CD-GAME}
  \centering
  \begin{tabular}{ccccccccccccc}
    \hline
Datasets&\multicolumn{2}{c}{\textbf{DBLP}}&\multicolumn{2}{c}{\textbf{Cora}}&\multicolumn{2}{c}{\textbf{Photos}}&\multicolumn{2}{c}{\textbf{Citeseer}}&\multicolumn{2}{c}{\textbf{Flickr}}&\multicolumn{2}{c}{\textbf{OGBN-Arxiv}}\\
\hline
       -      &M1&M2&M1&M2&M1&M2&M1&M2&M1&M2&M1&M2\\
       \hline
       \textbf{Clean} &3.61\%&34.11\%&3.25\%&53.55\%&6.91\%&50.77\%&5.37\%&58.32\%&4.63\%&47.35\% &3.88\%&43.97\%\\
       \hline
       \textbf{CD-ATTACK} &6.06\%&44.89\%&3.82\%&64.52\%&8.11\%&57.62\%&6.33\%&63.72\% &7.18\%&54.16\%&5.73\%&56.56\%\\
	\textbf{CD-DEFENSE} &3.58\%&35.67\%&3.32\%&55.73\%&7.02\%&51.62\%&5.47\%&60.02\% &4.87\%&50.17\% &3.94\%&47.06\%\\

    \textbf{CD-GAME} &4.54\%&41.02\%&3.58\%&61.02\%&7.69\%&56.83\%&5.83\%&62.31\%   &6.13\%&52.67\% &4.69\%&52.09\%  \\
  
  \hline
\end{tabular}
 \label{table:game}
\end{table*}

\begin{figure*}[!h]
\begin{center}
\includegraphics [width=1\textwidth]{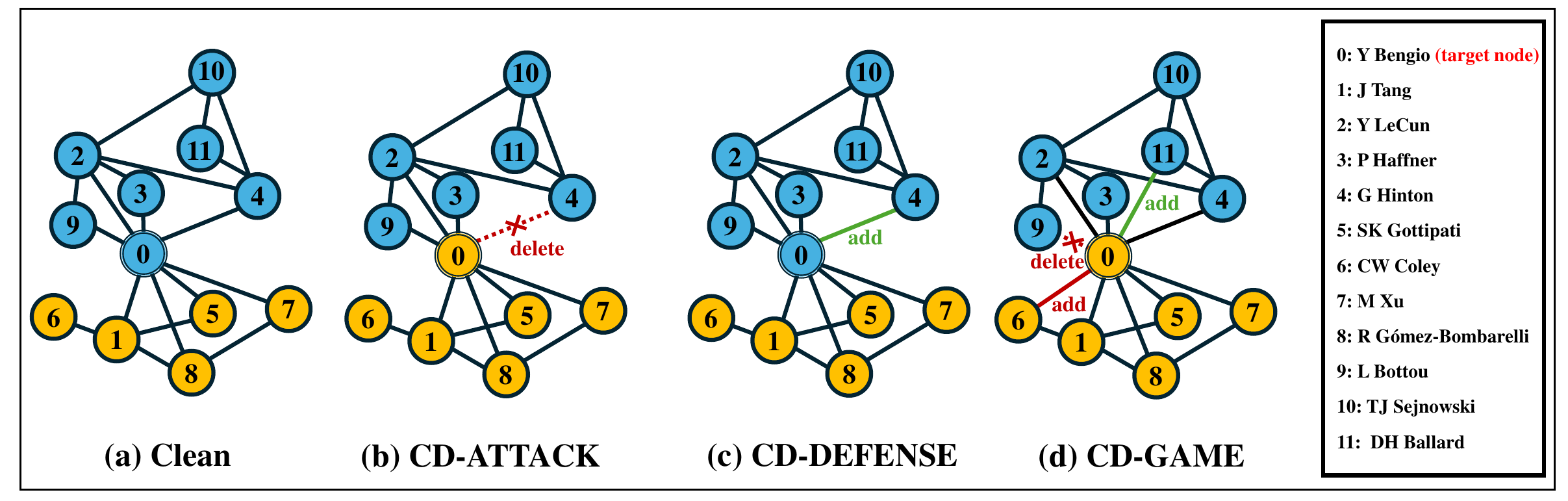}
\end{center}
\caption{A subgraph of DBLP to illustrate how CD-GAME affects decisions. The blue represents the machine learning community, and the yellow represents the Bioinformatics community. (a) Results of surrogate community detection model on original clean graph; (c) CD-ATTACK removes an edge and changes the community assignment of the target node; (c) CD-DEFENSE adds the same edge back to the original state; and (d) to bypass the defender, the attackers have to use a larger budget and perturb more camouflaged edges in CD-GAME.}
\label{fig.visualgame}
\vspace{-0.5cm}
\end{figure*}

\subsubsection{Baselines}
We use the following approaches as our defense baselines:
\begin{itemize}

\item \textbf{Inverse-DICE}. It first adds edges incident to $C^+$, then deletes edges between $C^+$ and the rest of the graph.

\item \textbf{Network Enhancement (NE)} \cite{wang2018network} is a graph denoising algorithm based on diffusion techniques. 

\item \textbf{Randomized Smoothing (RS)} \cite{jia2020certified} builds a smoothed function to ensure that the output unchanged when the noise is bounded.

\item \textbf{RobustECD-GA} \cite{zhou2021robustecd} combines modularity and the number of clusters in a fitness function to enhancement community structure.

\item \textbf{RobustECD-SE} \cite{zhou2021robustecd} Integrates multiple information of community structures captured by vertex similarities.
\end{itemize}

In this section, we initialize with the perturbed graph from CD-ATTACK and use various defenses to mitigate adversarial perturbation impacts. 

\subsubsection{Defense performance}

Table \ref{table:defense} lists the experimental results for the six datasets. Among all approaches, CD-DEFENSE achieves the best performance across all datasets in terms of both metrics, demonstrating its superiority. We observe that the defense effectiveness of RS is suboptimal. A reasonable explanation for this observation is that when the attack budget exceeds a certain threshold, the efficacy of RS diminishes rapidly. In addition, RobustECD-GA and NE perform worse than CD-DEFENSE due to its lack of consideration for node attributes. And RobustECD-SE is based on random sampling of similarities rather than the gradient of global graph similarities in CD-DEFENSE.

\subsubsection{Effect of Defense Budget}
We compare different defense approaches with varying defense budgets.  NE and RS do not support a defense budget, and the results are presented in Figures \ref{fig:budget3} and \ref{fig:budget4}. As we increase the defense budget, the performance of CD-DEFENSE improves, indicating that a larger budget is beneficial for successful defense. However, Inverse-DICE does not exhibit the same sensitivity to budget changes. It shows no defensive effect at smaller budgets.

\subsection{Interactive Attack-Defense Game}
In this section, we utilize CD-GAME to investigate the dynamics of interaction between the attacker and defender.

\subsubsection{Stackelberg Equilibrium}\label{graphgamesec}

Table \ref{table:game} presents the experimental results, which illustrate the strategic trade-offs inherent in our game-theoretic framework. The findings reveal that the highly effective perturbations generated by CD-ATTACK are also the ones most easily defended against by CD-DEFENSE.
Within CD-GAME, the attacker reaches an equilibrium state. Here, the attacker abandons the pursuit of optimal attack effectiveness and instead prioritizes enhanced stealth. It accounts for the defender’s potential decisions and adjusts its attack strategy accordingly, allowing a portion of its perturbations to bypass the defenses and achieve an attack effect. While CD-ATTACK’s strategy delivers the highest effectiveness, it is also the most likely to be detected and defended against. To bypass the defender, the attacker thus adopts strategies that are less effective but more camouflaged. 

\begin{figure*}[h]
    \centering
    \begin{subfigure}[b]{0.18\textwidth}
        \includegraphics[width=\textwidth]{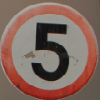}
        \caption{Raw image}
    \end{subfigure}
    \hfill 
    \begin{subfigure}[b]{0.18\textwidth}
        \includegraphics[width=\textwidth]{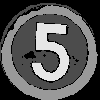}
        \caption{Clean}
    \end{subfigure}
    \hfill 
    \begin{subfigure}[b]{0.18\textwidth}
        \includegraphics[width=\textwidth]{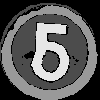}
        \caption{CD-ATTACK}
    \end{subfigure}
    \hfill 
    \begin{subfigure}[b]{0.18\textwidth}
        \includegraphics[width=\textwidth]{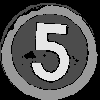}
        \caption{CD-DEFENSE}
    \end{subfigure}
    \hfill 
    \begin{subfigure}[b]{0.18\textwidth}
        \includegraphics[width=\textwidth]{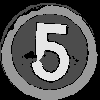}
        \caption{CD-GAME}
    \end{subfigure}
    \caption{The traffic sign segmentation results. (a) The raw traffic sign image; (b) Results of surrogate community detection model on raw clean traffic sign image; (c) CD-ATTACK changes the segmentation number to 6; (d) CD-DEFENSE recover the segmentation number to 5; and (e) in CD-GAME, attackers achieve sub-optimal results with larger budget cost.}
    \label{fig:is}
\vspace{-0.5cm}
\end{figure*}

\subsubsection{Visualization}

To better understand how CD-GAME works, we target a subgraph of DBLP consisting of 12 nodes, as shown in Figure \ref{fig.visualgame}. We set node $0$ as the target node, which corresponds to \emph{Yoshua Bengio}. A standard CD-ATTACK removes the most influential edge with largest similarity, but this obvious change is easily detected and reversed by the defender. In our CD-GAME scenario, the attacker and defender reach an equilibrium. Interestingly, in order to reduce the chance of being discovered by the defender, the attacker chooses to perturb camouflaged edges with less node similarity. In order to achieve the attack objective, the attacker has to spend a larger budget. In this example, the attacker uses two budgets (adding and deleting an edge) to successfully change the community assignment of the target node.
These disturbed edges are not the primary targets identified by the defenders. Therefore, the defender opts to add another edge that does not counteract the attacker's decision. This example shows the interaction between attacker and defender. A more powerful strategy is usually easily exposed to defense. As a result, if the attacker ignores the defender's influence, the attacker's actions are likely to be discovered and countered. We find that the attacker bypasses the defender's defense by adopting some less effective but more camouflaged strategies. Meanwhile, attackers need to pay a higher cost if they want to get better results, meaning a larger budget is needed.

\subsubsection{Generalization of Detection Model}
To validate the generalization capability of our methods, we evaluate them on two classic community detection models, Louvain~\cite{rosvall2008maps} and Leiden~\cite{traag2019louvain} algorithms, on DBLP. The results are summarized in Table~\ref{table:Generalization}. The performance is consistent with that on the surrogate detection model, which indicates that our approach is not overfit to a specific detection model and can be effectively generalized to other unseen models.
\begin{table}[t]
  \caption{Performance on different detection models}
  \centering
  \begin{tabular}{ccccc}
    \hline
Models&\multicolumn{2}{c}{\textbf{Louvain}}&\multicolumn{2}{c}{\textbf{Leiden}}\\
\hline
       -      &M1&M2&M1&M2\\
       \hline
       \textbf{Clean} &4.02\%&37.45\%&4.21\% &37.18\%\\
    \hline
	\textbf{CD-ATTACK} &5.86\%&45.19\% &5.63\% &45.42\%\\
	\textbf{CD-DEFENSE} &3.88\%&36.71\%&3.65\% &36.94\%\\
	\textbf{CD-GAME} &4.83\%&42.34\%&4.59\% &42.61\%\\
  \hline
\end{tabular}
 \label{table:Generalization}
\end{table}

\subsection{A Case Study: Traffic Sign Segmentation}

Following the normalized cut approach \cite{shi2000normalized}, we showcase a traffic sign image by formulating community detection as the image segmentation task. We represent each pixel of the RGB image as a node within a graph. The pixel's 3-channel color values serve as its descriptive features. The adjacency matrix is constructed as follows:
\begin{align}
A_{ij} = \left\{
    \begin{array}{ll}
    1 
    & \mathrm{if} \|X_{i}-X_{j}\|_2 < r~\mathrm{and}~\|F_{i}-F_{j}\|_2 < \alpha \\
    0 & \mathrm{otherwise}
    \end{array}
    \right.
\end{align}
where $X_i$ and $F_i$ respectively denote the spatial location and features of pixel $i$, and $r,\alpha$ are predefined thresholds.

We take a traffic sign to conduct the image segmentation experiment. In this task, our goal is not to have pixels of $C^+$ spread out across the communities. In contrast, the attack goal is to change the segmentation number of the traffic sign from 5 to 6.  We select the patch near the tail of number 5 as the target nodes $C^+$. In this task, we choose $r=1$ and $\alpha=20$.   Figure \ref{fig:is} visually illustrates the original images and the  segmentation results of different methods.  We find that the surrogate community detection model achieves good performance on clean traffic sign image. And CD-ATTACK changes the segmentation number from 5 to 6. CD-DEFENSE recovers the segmentation number to 5. In CD-GAME, after reaching the Stackelberg equilibrium, the attackers are hard to achieve a similar performance as in CD-ATTACK even with larger budget.

\section{Related Work}\label{sec:relatedwork}

\noindent\textbf{Self-supervised GNN}. 
Self-supervised Graph Neural Networks (GNNs) learn from unlabeled data using objectives that are broadly categorized as contrastive or generative. Contrastive methods align multi-view representations by maximizing mutual information, with research focusing on data augmentation and negative sampling. Techniques include corruption (DGI~\cite{velivckovic2018deep}, DCI~\cite{wang2021decoupling}), graph diffusion (MVGRL~\cite{hassani2020contrastive}), masking (GRACE~\cite{zhu2020grace}, GCA~\cite{zhu2021gca}), latent noise (COSTA~\cite{zhang2022costa}), and spectral augmentation (SFA~\cite{zhang2023spectral}), alongside negative-sample-free approaches like BGRL~\cite{thakoor2022bgrl} and CCA-SSG~\cite{zhang2021ccassg}. Generative methods, often autoencoder-based, learn by reconstructing information. Early work like VGAE~\cite{kipf2016vgae} focused on structural reconstruction, while others used a perturb-and-predict strategy on features. More recently, focus has shifted to decoder design, seen in GraphMAE~\cite{hou2022graphmae, hou2023graphmae2} re-masking, WGDN~\cite{cheng2023wiener} spectral decoder, and S2GAE~\cite{tan2023s2gae} cross-correlation decoding.

\noindent\textbf{Attacks on Community Detection}. 
Attacks on community detection are mainly categorized as target node attacks~\cite{nagaraja2010impact,waniek2018hiding,fionda2017community,chen2020multiscale,chen2019ga}, which hide a specific node, or target community attacks~\cite{fionda2022community,fionda2022community2,zhao2023self,liu2019rem,fionda2021community,fionda2017community}, which obscure an entire community. This work focuses on black-box target node attacks. Previous studies proposed heuristic solutions based on degree centrality~\cite{nagaraja2010impact}, ROAM and DICE algorithms~\cite{waniek2018hiding}, and modularity-based deception~\cite{fionda2017community}. While we adopt the deception quantification measures from~\cite{waniek2018hiding}, our method, CD-ATTACK, differs significantly. It is designed to (1) target modern deep graph-based community detection methods and (2) handle attributed graphs, a capability lacking in prior work.

\noindent\textbf{Defenses on Community Detection}. 
Adversarial defense for community detection is a sparsely studied field. The only dedicated work, \cite{jia2020certified}, uses randomized smoothing \cite{cohen2019certified} to certify robustness against structural perturbations. However, its limitation is that it treats graphs simply as edge sets, making its effectiveness across different graph types unclear. A related approach is network enhancement \cite{wang2018network}, which recovers the graph structure using diffusion algorithms before running detection. RobustECD \cite{zhou2021robustecd} is also a network enhancement approach based on  genetic algorithm and node similarity.

\noindent\textbf{Game Theory on Community Detection}. 
The learning dynamics of attack and defense in community detection are currently unexplored. Traditional game-theoretical approaches model the community formation process \cite{moscato2019community, jonnalagadda2016survey, chopade2016framework} but neglect the adversarial security game. While a recent study \cite{miller2023using} uses game theory to select from existing attack and defense strategies, our work differs fundamentally. We focus on the security game itself and aim to model the interactive dynamics between an attacker and a defender, rather than simply choosing established methods.

\section{CONCLUSION}\label{sec.con}

This paper tackles adversarial attack and defense for community detection. We propose CD-ATTACK, a scalable attack method, and CD-DEFENSE, an unsupervised defense. We also introduce CD-GAME, a game-theoretic framework to model the interactive dynamics between them and provide valuable insights. Experiments on six real-world datasets show our methods, built upon a novel GNN-based detection model, significantly outperform competitors.

\ifCLASSOPTIONcaptionsoff
  \newpage
\fi



%



\printbibliography 

%

\vspace{-1.5CM}

\begin{IEEEbiography}[{\includegraphics[width=1in,height=1.25in,clip,keepaspectratio]{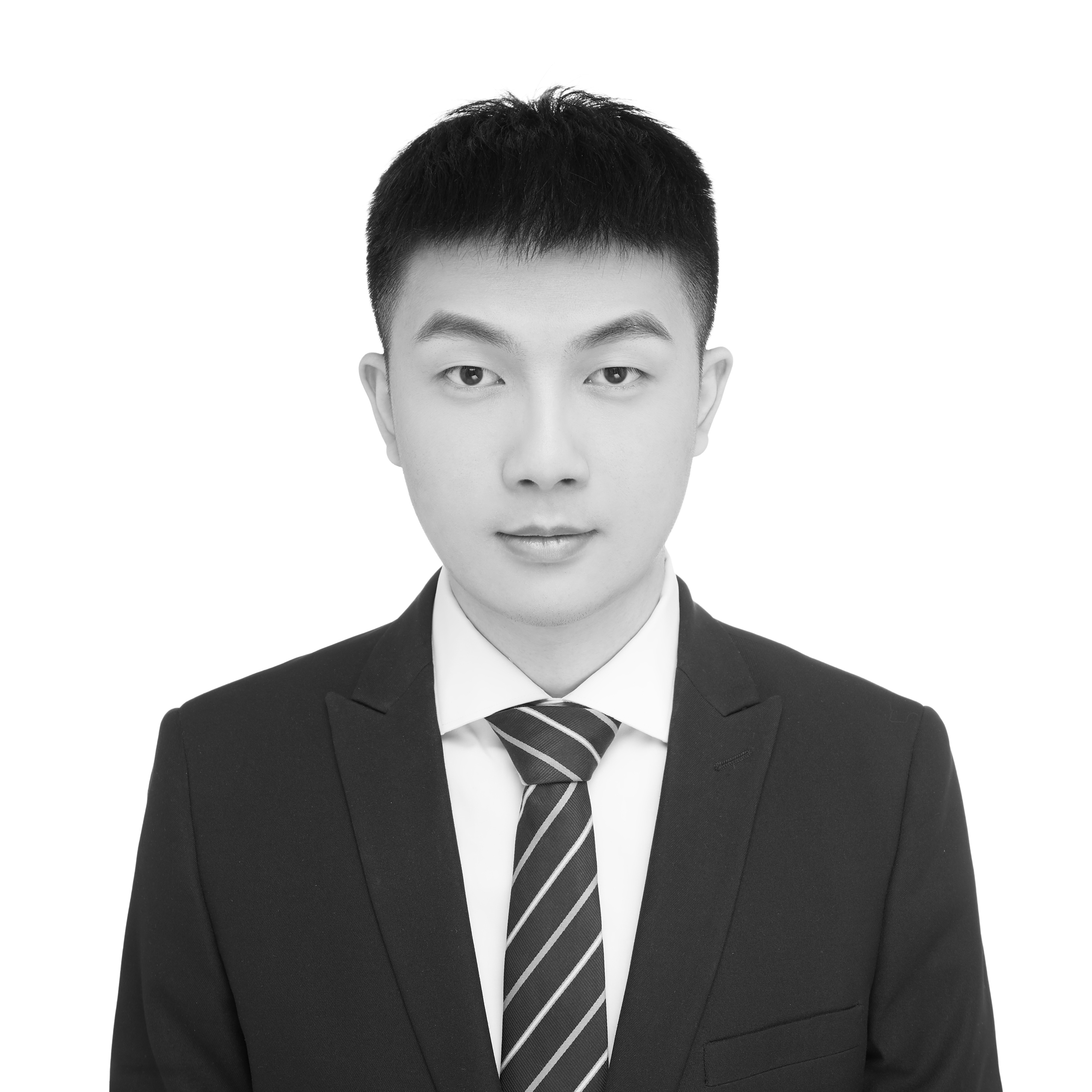}}]{Yifan Niu}
is a Ph.D. student at The Hong Kong University of Science and Technology. He received the bachelor's degree and the master's degree at Beijing University of Posts and Telecommunications. His research interests lie in machine learning on graph data, large language model and computational biology.  He has published several papers in top conferences such as ICLR, ICML, ACL and KDD.
\end{IEEEbiography} 

\vspace{-2CM}

\begin{IEEEbiography}[{\includegraphics[width=1in,height=1.25in,clip,keepaspectratio]{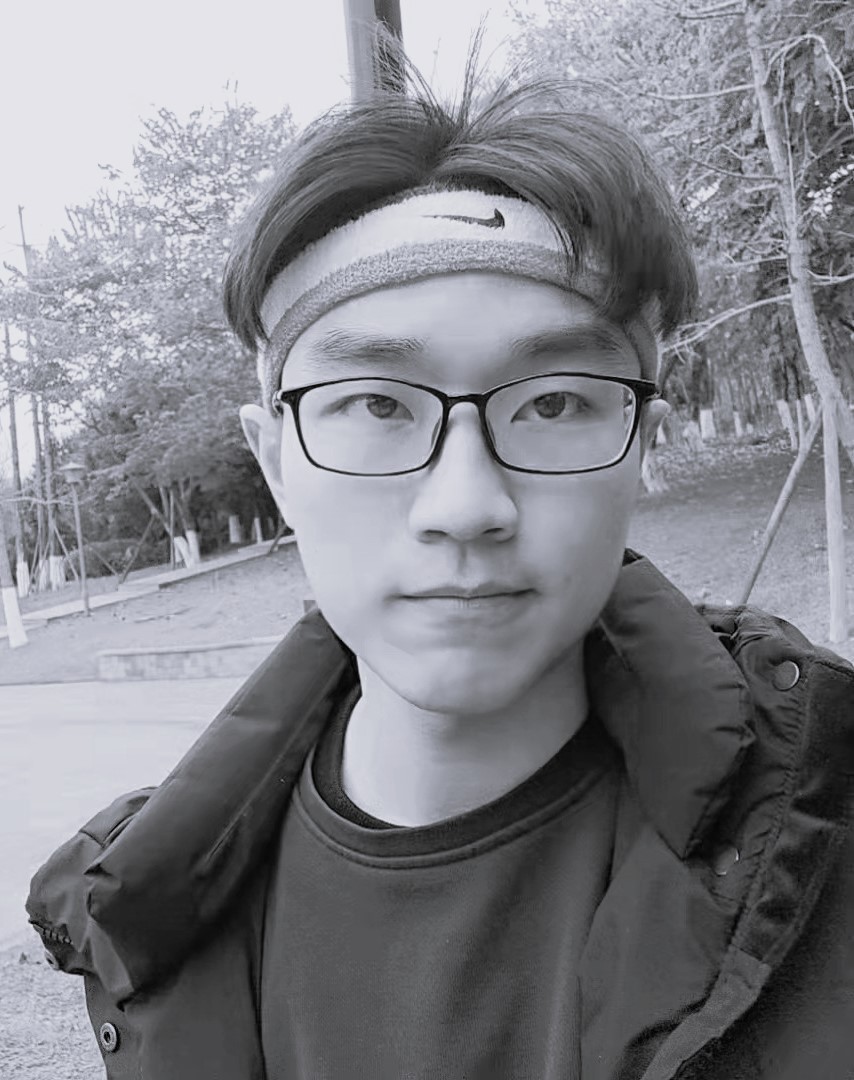}}]{Aochuan Chen} is a Ph.D. student at The Hong Kong University of Science and Technology. He received the bachelor's degree at Tsinghua University in 2022. His research interests lie in machine learning, optimization, and graph learning. He has published several papers as the leading author in top conferences such as CVPR, NeurIPS, ICLR, ICML and KDD.
\end{IEEEbiography}

\vspace{-2CM}

\begin{IEEEbiography}[{\includegraphics[width=1in,height=1.25in,clip,keepaspectratio]{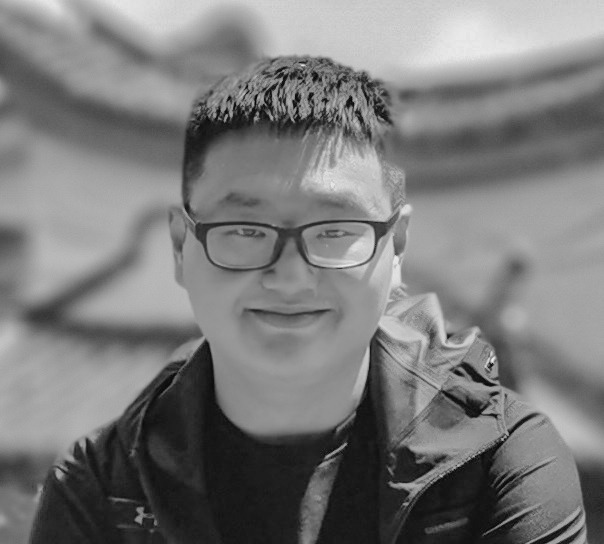}}]{Tingyang Xu} 
is a Senior Researcher at DAMO Academy, Alibaba Group and Hupan Lab, earned his Ph.D. from The University of Connecticut. His research encompasses deep learning applications for de novo drug design, molecular property prediction, and molecular dynamics simulation. His work has been published in top-tier conferences, including NeurIPS, ICML, SIGKDD, AAAI and VLDB. 
\end{IEEEbiography}

\vspace{-2CM}

\begin{IEEEbiography}[{\includegraphics[width=1in,height=1.25in,clip,keepaspectratio]{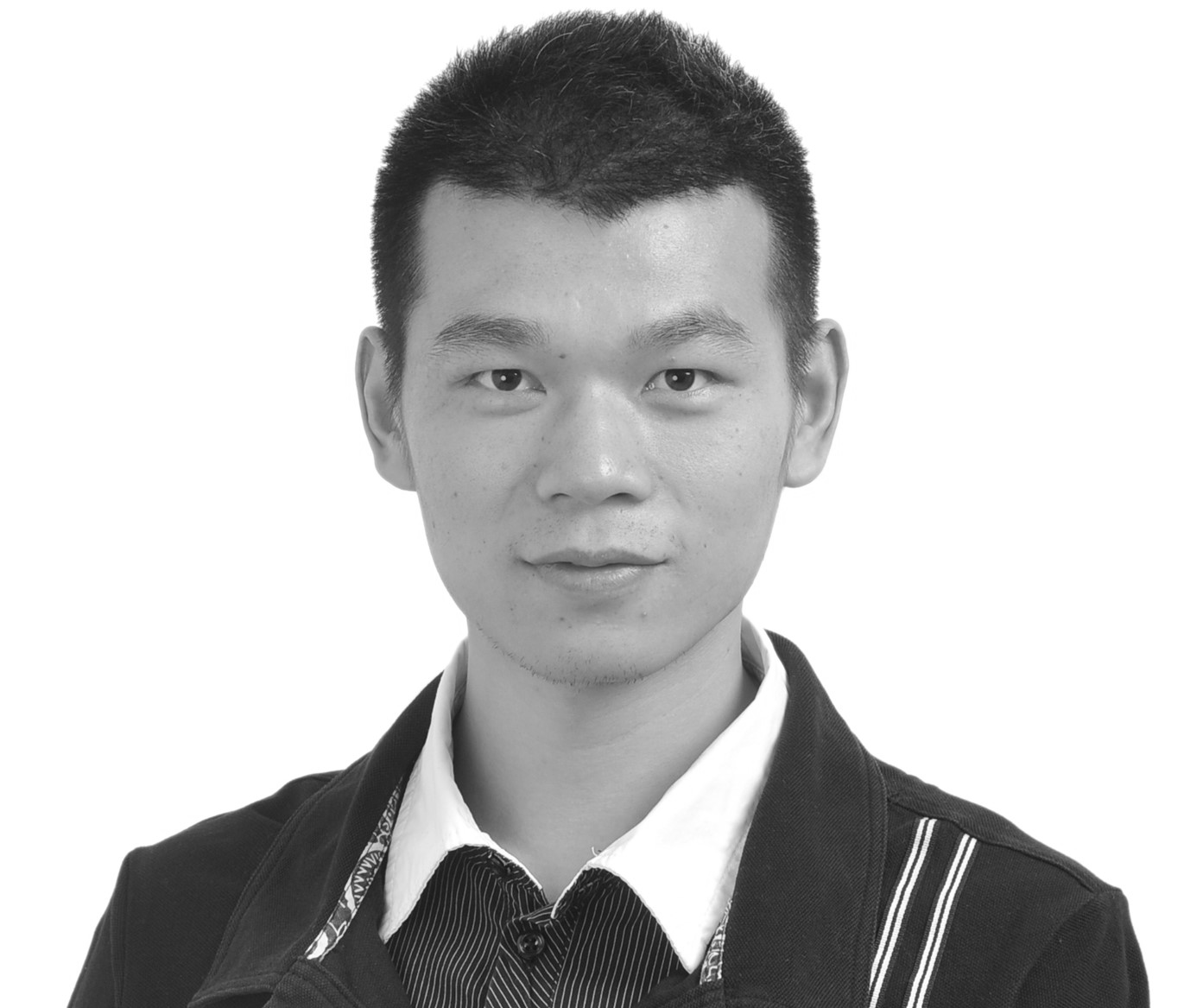}}]{Jia Li}
 is an assistant professor with the Data Science and Analytics Thrust, Information Hub, The Hong Kong University of Science and Technology. He received the Ph.D. degree at The Chinese University of Hong Kong in 2021. His research interests
include machine learning, data mining and deep graph learning. He published several papers as the leading/corresponding author in top conferences such as T-PAMI, KDD, NeurIPS and ICML. He received KDD 2023 Best Paper Award.

\end{IEEEbiography}



\end{document}